\pdfoutput=1
\documentclass[11pt]{article}

\usepackage[]{naacl2021}

\usepackage{times}
\usepackage{latexsym}
\usepackage{url}
\usepackage{enumitem}

\usepackage{amsmath}
\usepackage{amssymb}
\usepackage{graphicx}
\usepackage{hhline}
\newcommand{\R}{\mathbb{R}}

\newcommand{\B}{\fontseries{b}\selectfont}

\usepackage[T1]{fontenc}
\usepackage[utf8]{inputenc}

\usepackage{microtype}

\title{Intent Features for Rich Natural Language Understanding}

\author{Brian Lester$^\spadesuit$\thanks{$^*$Now an AI Resident at Google} \and Sagnik Ray Choudhury$^\diamondsuit$ \thanks{$^*$Work done while at Interactions} \and Rashmi Prasad$^\spadesuit$ \and Srinivas Bangalore$^\spadesuit$ \\
  $^\spadesuit$Interactions, 41 Spring Street, New Providence, NJ 07974 \\
  $^\diamondsuit$University of Copenhagen, Denmark \\
  \texttt{\{blester,rprasad,sbangalore\}@interactions.com, src@di.ku.dk}
}

\begin{document}
\maketitle
\begin{abstract}
Complex natural language understanding modules in dialog systems have a richer understanding of user utterances, and thus are critical in providing a better user experience. However, these models are often created from scratch, for specific clients and use cases, and require the annotation of large datasets. This encourages the sharing of annotated data across multiple clients. To facilitate this we introduce the idea of \textit{intent features}: domain and topic agnostic properties of intents that can be learned from the syntactic cues only, and hence can be shared. We introduce a new neural network architecture, the Global-Local model, that shows significant improvement over strong baselines for identifying these features in a deployed, multi-intent natural language understanding module, and, more generally, in a classification setting where a part of an utterance has to be classified utilizing the whole context.
\end{abstract}

\section{Introduction}

While generic dialog systems, or chatbots, such as Amazon Alexa or Google Assistant, are increasingly popular, to date, most industrial dialog systems are built for specific clients and use cases. Typically, these systems have the following: 1.\ A natural language understanding (NLU) module to analyze the user utterance, 2.\ A dialog manager module to reason over the analyzed utterance and decide on an action, and 3.\ A natural language generation module to generate an appropriate response based on the action. 

Typically, an NLU module has two purposes: understanding the intent or goal of an utterance (classification) and identifying the entities in the utterance (slot filling). As dialog managers have evolved from simple flow-based systems to information state update systems \cite{traumInformationStateApproach2003}, NLU modules have progressed past simple single intent detection and flat slot filling to multiple intents and nested entities \cite{8639605}. As these dialog systems need to be rebuilt for each client, the NLU module faces a significant data bottleneck; it is time-consuming and expensive to collect data, develop a domain-specific annotation scheme, and annotate data. Therefore, it is imperative that the data is shared across clients as much as possible.

In a production dialogue system, there are often similar situations that require drastically different responses. For example, ``I want to cancel my subscription.'' and ``I am thinking about canceling my subscription.'' are very similar. They are both about the canceling of a subscription. However, they differ in the users conviction. The latter user is much more likely to not cancel if offered a discount. Making this distinction is critical for creating sophisticated and nuanced dialogue systems. A common approach to solve this problem would be to split the intent space so the dialogue manager can differentiate between these examples, creating a \texttt{cancel} and a \texttt{think-cancel} intent. Using intents to recognize specific situations leads to data sparsity as each intent is broken into many sub-categories like present vs.\ past tense, how certain a user is in their actions, and if the user has tried an action or not. There would be very few examples of each intent. Additionally, the combinations of different sub-categories would cause a combinatorial explosion of intents. Another short-coming of fine-grained intents is the loss of compositionality. Fundamentally the \texttt{cancel} and \texttt{think-cancel} intents are very similar, but because they are modeled as independent output classes, there is not a shared representation of these labels the model can lean on.

In order to avoid these shortcomings, and allow for many examples per intent, we factor out these small differences in situations into what we call intent features. Intent features are a set of domain-independent properties for intents that can primarily be understood from the syntax of the utterance. These intent features represent specifics of situation, such as tense, without having a massive intent space. By decoupling these small differences, we can keep the intent categories general, while still providing the dialogue manager with the information it needs for nuanced, human-like responses.

In a multi-intent setting where each clause in the utterance has an intent, intent features reduce to the problem of classification of a span embedded within a larger utterance. We propose a new model, the Global-Local model, for this problem which shows significant improvement over strong baselines.

\section{Intent Features}
\label{sec:intent-feature-def}

\begin{table*}[h]
\begin{tabular}{l|l|l|l|l|l|l|l}
text                   & topic/intent & attr-cf & attr-ev & cf     & modality   & negation & tense   \\ 
\hhline{=|=|=|=|=|=|=|=}
I am trying to install & install      & self    & self    & inform & modal-try  & positive & present \\
and                    & -            & -       & -       &        & -          & -        & -       \\ 
I see a problem        & general      & self    & self    & issue  & other      & positive & present \\
\end{tabular}
    \caption{
        A sample utterance from our dataset with multiple intents and features. Each row represents an intent span and the columns are the features that apply to that particular intent. We see that intents are general categories of actions like ``install'', while intent features yield specifics of the current state of the user. Given the ``modality'' and ``tense'' features, we see that the user is currently in the middle of installing the program, rather than telling us they installed it last week. ``cf'' stands for communicative function.
    }
    \label{tab:sample-utterance}
\end{table*}

Table \ref{tab:sample-utterance} shows a sample utterance with its intents and features. This is a multi-intent setting where non-overlapping spans of an utterance have different intents. Each intent span has the following features:

\textbf{Communicative functions}: The communicative functions (cf) captures what kind of response (or action) the user is trying to elicit from the system. We define five such functions:
\begin{itemize}[leftmargin=*]
\setlength\itemsep{0em}
\item \texttt{inform}: The user is informing the system about something. Typically, these intents are a response to a question or they represent background information surrounding the main purpose of the utterance. For example, in the utterance, ``I am installing X but it keeps saying I have an error'', the first clause has a communicative function of \texttt{inform}. The user provides background information about installing something on a device and then presents a problem with the install procedure, which would have a communicative function of \texttt{issue}.
\item \texttt{issue}: The user is saying that something has gone against their expectations (see above for an example).
\item \texttt{request-action}: The user requests for some action to be undertaken in response to the request, or requests help with something. For example, ``I would like to install X.'' 
\item \texttt{request-confirm}: The user is requesting confirmation, or disconfirmation, of their belief. Often this warrants a yes/no answer. For example, one expects a yes or no from, ``Was my installation successful?''
\item \texttt{request-info}: The user is requesting some information about something. These are typically expressed as ``wh/how'' questions, such as: ``How can I install X?''
\end{itemize}

All of our running examples above share the intent of installing software; however, differences in phrasing warrants different responses. An \texttt{inform} does not typically require a targeted reply from the system, whereas for an \texttt{issue}, the system should start the response with ``I am sorry you are having trouble.'' 

\textbf{Attribution}: Attribution is concerned with \textit{agency}. There are two types of attribution. The first type is the of attribution of the communicative function (\textbf{attr-cf}) and it deals with who is the primary source of the content of the topic. The second type is the attribution of the event/action (\textbf{attr-ev}) of a topic and describes who is the agent of the event or action. This is perhaps best elucidated by an example. In Table~\ref{tab:attribution}, we see multiple utterances that all have the intent \texttt{payment}, but we can see how the attribution features change as both the payer and the informer of the payment change. Both \textbf{attr-cf} and \textbf{attr-ev} take values \texttt{self} (when the agent is the user) and \texttt{other}.   

\begin{table*}[]
    \centering
    \begin{tabular}{l | r r}
        Utterance & Attribution CF & Attribution Ev \\
        \hhline{=|==}
        I have paid \$\$ & self & self \\
        I got an email confirming I paid \$\$ & other & self \\
        I was charged  \$\$ & self & other \\
        I got an email confirming that I was charged \$\$ & other & other \\
    \end{tabular}
    \caption{
        Different types of attribution with the same \texttt{payment} intent. The dialogue manager would react differently depending on whether the user paid voluntarily vs she was charged or if she was only informed that she was charged.
    }
    \label{tab:attribution}
\end{table*}

\textbf{Negation}: Topics of many intents are represented in their negated versions, as well. For example, in the software domain, the \texttt{compatibility} intent models whether a piece of software is compatible with some device. A negation feature would denote incompatibility. The negation feature takes values \texttt{positive} and \texttt{negative}.

\textbf{Tense}: Events and actions can occur in the past, present, or future, which is modeled by the tense feature using values of \texttt{past}, \texttt{present}, or \texttt{future}. The steps to solve a problem as it occurs are often quick-fixes,
whereas the first step when fixing a problem that occurred in the past is often information gathering. The tense feature allows the dialogue manager to distinguish between these two possibilities. Tense information is common in the annotation of event extraction, such as in ACE 2005 dataset \cite{ace}.

\textbf{Modality}: The real-world actions and events represented by an intent can also be viewed in terms of a modality of certainty, that is, whether or not the event or action actually occurred, and to what degree. We consider two types of modality. The first is \textit{possibility}---the expression of the event as hypothetical, or being possible, rather than certain, as in, ``I am \textit{planning/going} to install X on my laptop.'' We also consider \textit{attempts at action}. An expression can imply that it is unclear whether the action was completed or is in the attempted stage. This is expressed with modifying verbs, such as, ``try'', as in, ``I am \textit{trying} to install X.'' This feature takes the values \texttt{modal-poss}, \texttt{modal-try}, and \texttt{other}. A version of Modality is present in event extraction datasets like ACE 2005 \cite{ace}, but instead of just marking an event as ``Asserted'' or ``Other'', our version of Modality distinguishes between different aspects of hypothetical events.

\section{Modeling}

There are four different model types we explored for intent features that we detail below. However, before we can annotate an intent with a feature, we need to have an intent span. First, we describe our intent span extraction model whose predictions are used as intent spans. 

\subsection{Multi-Intent as Annotatable Spans}
The intents in our system are often conditionally dependent. Some intents even appear sequentially, for example, the \texttt{cancel} intent is often followed by the \texttt{refund} intent, as users tend to request a cancellation first and then ask for a refund. Therefore, we modeled our multi-intent system as a sequence tagging problem, where intent spans are encoded as token level annotations with the IOBES tagging scheme \cite{ratinovDesignChallengesMisconceptions2009}. We used a standard BiLSTM-CRF architecture following \citet{maEndtoendSequenceLabeling2016}. Each input token is represented both as a character composition, by running a small convolutional neural network with a filter size of $3$ over the characters and doing max-over-time pooling as in \citet{DosSantos:2014:LCR:3044805.3045095:14}, and as a word embedding. We use the concatenation of multiple word embeddings, GloVe embeddings \cite{penningtonGloVeGlobalVectors2014}, as well as $100$ dimensional, in-domain embeddings trained in-house, following \citet{lester-2020-multiple}. The token sequence is then fed into an bidirectional LSTM \cite{10.5555/1986079.1986220}, where the LSTM \cite{10.1162/neco.1997.9.8.1735} in each direction has a size of $200$, and projected to the final label space. Finally a Conditional Random Field (CRF) \cite{10.5555/645530.655813} with constrained decoding \cite{lester-etal-2020-constrained} is used to produce the final sequence of intents. This model was trained using SGD with momentum using $0.0015$ as the learning rate, $0.9$ for momentum, and a batch size of $10$. Model results were satisfactory, but not the focus of this paper. Instead, intent spans are the atomic unit of text that can be annotated with intent features and can be used as features for a downstream intent feature model.

\subsubsection{Convolutional Baseline}
The first approach was to assume that the feature labels for an intent are local to that intent span, and, therefore, each intent span can be fed into a classifier independently of the other intent spans. Under this assumption, we used a convolutional neural network with parallel filters \cite{kimConvolutionalNeuralNetworks2014}, as it is a strong baseline used in several of our production systems. We used parallel filters of size $3$, $4$, and $5$ with $100$ filters each. Max-over-time pooling was used to produce a final span representation, which is projected into the label space. This model was trained using Adadelta \cite{Zeiler2012ADADELTAAA} with an initial learning rate of $1.0$ and a batch size of $50$. However, this approach misses possible dependencies across spans. Some features (such as ``tense'') are naturally co-dependent among spans; the use of a past tense verb in one span dictates that all spans in the utterance are past tense, even when there is no explicit signal from the span itself. While less intuitive, the ``communicative function'' features are conditional as well: an utterance such as, ``I would like to order a pizza, but I am having a problem'' (a \texttt{request-action} followed by an \texttt{issue}) is far more common than an utterance like ``I am having a problem, I would like to order a pizza'' (an \texttt{issue} followed by a \texttt{request-action}). It follows that the ``independence of intent spans'' assumption will become problematic and a contextual model that takes other spans into account will be needed.

\subsubsection{Contextual Features with a BiLSTM-CRF}
This motivated us to reuse the BiLSTM-CRF architecture we used for intents for the intent features, as well. This model takes the utterance as input, just like the intent model. This approach has a potential pitfall, the intent model and the feature model may produce different boundaries which need to be heuristically merged. A small modification to this approach is to use a cascading tagger where the output of the intent tagger is used in the input to the feature tagger. This is done by creating an embedding that represents the span each token is within and concatenating it to the token representation. This gives the feature tagger information about the span boundaries and should keep the spans synced between the intent and feature models. However, the actual intent labels need to be masked. Instead of seeing \texttt{intent=issue} as a feature, the feature model will just see \texttt{intent}. This is required because we want the feature labels to be reusable and therefore unconditioned on exact intent label. Intent features are applied to intent spans within an utterance, meaning our BiLSTM-CRF tagger is a natural baseline that considers the global context of an utterance.

\subsubsection{Global-Local Model}

\begin{figure}
    \centering
    \includegraphics[width=0.45\textwidth]{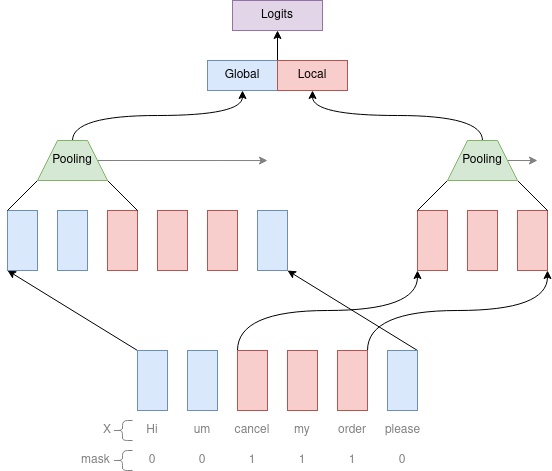}
    \caption{The architecture of our Global-Local model. There are four distinct phases of the model. First, the input is encoded into a sequence of vector representations. This can be as simple as word embeddings or it can use a more complex encoding like a BiLSTM. Then, the local span is extracted from the sequence using the input mask. Global and Local pooling functions are applied to create two vectors, which are joined by concatenation. The local vector encodes the features of the span while the global vector encodes the features of the span as contextualized by the whole input. Finally, this joint representation is used for classification.}
    \label{fig:arch}
\end{figure}

Our fourth approach is a new model architecture we call the Global-Local model. This model aims to create a targeted representation for a subsection of an utterance while also infusing information derived from the whole utterance. An utterance $U$ of $n$ tokens and a subsequence of $k$ tokens from $U$, are first encoded into matrices of dimension $n \text{ x } e$ and $k \text{ x } e$, respectively, where $e$ is the dimension of some shared embedding space. This encoding can be as simple as word embeddings or more complex like a BiLSTM encoder. A ``global'' pooling function $g: \R^{n \text{ x } e} \mapsto \R^{e}$ then collapses the global sentence matrix to a sentence vector and another ``local'' pooling function $l: \R^{k \text{ x } e} \mapsto \R^{e}$ reduces the span matrix to a span vector (both with dimension $e$). The local vector is a representation based solely on the span, while the global vector is a representation of the span that takes the whole utterance into account. These vectors are concatenated to create the final representation for the span $S$. This representation is then projected into the output space. The pooling functions can be as simple as max or mean pooling, or as complicated as self-attention \cite{vaswaniAttentionAllYou2017}. Each example is represented as a sequence of tokens and a mask. The mask is a sequence of zeros and ones, aligned to the tokens, that marks a token as part of the local span (a one) or not (a zero). A diagram of the model architecture can be found in Figure~\ref{fig:arch}.

Our implementation uses lookup-table based word embeddings, the same embeddings used in our convolutional baseline, to create a sequence of vectors representing the input. Then a convolutional neural network with multiple parallel filters, followed by max-over-time pooling, is used as both the local and global pooling functions. We found that when $g$ and $l$ share parameters, results were a bit worse compared to when they are learned separately. Like our convolutional baseline, we use filter sizes of $3$, $4$, and $5$ with $100$ filters each. This model was trained with a cross-entropy loss using the Adadelta optimizer with an initial learning rate of $1.0$ and a batch size of $50$.

\section{Dataset}

The data consists of customer utterances. They were collected from the first customer turn in web-chat conversations between customers and agents from a software company after filtering out low content first turns such as ``Hi'', ``Hello'', and ``Hey''. Our training, validation, and testing datasets have $36{,}725$; $9{,}256$; and $4{,}993$ examples respectively. The data was annotated by a team of six (non-overlapping) commercial annotators over a period of a month and then corrected by an expert annotator. A small subset of the data was annotated (before the error correction) by two expert annotators. The agreement was $53\%$ between two expert annotators and $42\%$ between one expert and the other non-expert annotators. 

\section{Experiments}

\begin{table*}[]
    \centering
    \begin{tabular}{l| r r r r}
%                & \multicolumn{3}{c}{F1 Score for Intent Features} \\
        %\cline{2-4}
        Feature & BiLSTM-CRF &  \shortstack[r]{BiLSTM-CRF \\ Cascaded Tagger} & \shortstack[r]{ \\ Span-level \\ Convolutional} & Global-Local \\
        \hhline{=|====}
         Attribution CF         & 78.63 & 91.90 & 95.37 & \B{97.69} \\
         Attribution EV         & 80.06 & 92.27 & 95.86 & \B{98.16} \\
         Communicative Function & 69.07 & 89.22 & 90.12 & \B{91.92} \\
         Modality               & 79.31 & 92.61 & 96.60 & \B{99.36} \\
         Tense                  & 73.49 & 86.01 & 89.31 & \B{92.59} \\
         Negation               & 78.47 & 94.45 & 95.86 & \B{98.73} \\
    \end{tabular}
    \caption{
        F1 score of intent features using various models. BiLSTM-CRF is the feature tagger that was not given intent boundaries. Cascaded Tagger is the same BiLSTM-CRF model, except the intent boundaries are fed into the model. Span-level Convolutional is our model that classifies each intent span independently, and Global-Local is our new model that encodes both the span and a global view of the sentence. We see that our Global-Local model shows consistent improvements over other model types.
    }
    \label{tab:intent-features}
\end{table*}

The F1 scores for these models are reported in Table \ref{tab:intent-features}. The BiLSTM-CRF tagger without any information about the intent boundaries has the lowest performance. Our analysis suggests that it is difficult for the tagger to learn the span boundaries for the features. When that information is supplied---as seen in the cascaded tagger column---the results improve by a large margin. The span-level convolutional model, which is agnostic to the tokens of the other spans, performs much worse than the Global-Local model, which clearly validates our hypothesis that global information is valuable.  

\begin{table*}[]
    \centering
    \begin{tabular}{l|r r r r r r}
        %      & \multicolumn{6}{c}{F1 Score for Intent Features} \\
        Model & Attribution CF & Attribution Ev & CF & Modality & Tense & Negation \\
        \hhline{=|======}
        Global-Local              & \B{97.69} & \B{98.16} & \B{91.91} & \B{99.36} & \B{92.59} & \B{98.73} \\
        \quad -- Global Context   & 93.55 & 95.47 & 90.14 & 96.74 & 87.18 & 95.74 \\
        \quad -- Shared Embedding & 97.65 & 96.63 & 91.43 & 98.34 & 90.17 & 96.36 \\
    \end{tabular}
    \caption{
        Ablation of the Global-Local model. We see that removing the global context causes a large degradation in F1 score, implying that the strong performance of the Global-Local model is due to the global feature, not just the increased parameter count. We also see the removing the shared embedding hurts model performance but to a much smaller degree.
    }
    \label{tab:global-local-ablation}
\end{table*}

We further ablate the Global-Local model to understand the reasons for the performance gain in Table~\ref{tab:global-local-ablation}. To test if the performance improvement is only due to the larger parameter count, and not the global cues, we use \textit{only} the span as the input (as opposed to \textit{both} the utterance and the span), but the same Global-Local Model. If the Global-Local model is only stronger because it is larger, we should not see a drop in performance. As we can see in the ``-- Global Context'' row, limiting the model to only see the span causes large performance drops across the board. This model is even worse than the simple convolutional model. This implies that the global context is critical.% to performance.

The current implementation has a shared encoder step where the entire utterance in encoded into a sequence of vectors before the span is extracted and processed by the local pooling function separately. Doing this efficiently in a batched computing environment, like TensorFlow \cite{tensorflow2015-whitepaper}, is slightly tricky to implement. A much simpler model would feed the global utterance and the span separately, to be encoded and processed independently. Our ablations in the ``-- Shared Embedding'' row of Table~\ref{tab:global-local-ablation} shows that using a shared embedding space does yield performance gains, but it can be removed for the sake of easier model deployment and still maintain superior performance over the span-level model.

All models were trained with Mead-Baseline \cite{Baseline:2018}, an open-source library for the development, training, and export for deep neural networks for NLP.

\section{Deployment}

We have deployed a NLU component of a task-oriented, production dialogue system that produces intent features. The dialogue system deals with customer service in the retail software domain. The dialogue manager currently makes use of several intent features. The easier feature to use is negation and it is critical to understand user intent. It also uses the tense feature to understand if it needs to wait because a user is currently performing an action or if it can ask about the result because the action had already been performed. The next feature the dialogue manager plans to leverage is the modality features. Understanding the user's convection in an action, like canceling, can help make decisions about whether an upsale or discount would be effective.

In designing these intent features, we hoped they would be general enough to be transferable across domains without retraining a model on the new domain. Recent work with a new client in the general retail domain gave the opportunity for a small scale test. We were given approximately 500 sample utterances that had been annotated with general labels like, ``Is this utterance equivalent to an FAQ?'' This is very similar to our \texttt{request-info} intent feature. We ran our intent feature model on this new data and compared how many FAQ questions were labeled with \texttt{request-info}. We found that our model had high precision, $83.3\%$ of \texttt{request-info} utterances were in fact FAQ questions, but had low recall, only $40.5\%$. This small scale experiment suggests that our intent features are general, but the low recall means our specific model is probably overfit to the lexical features in our original domain.

\section{Previous Work}

Most popular intent taxonomies such as ATIS \cite{10.3115/116580.116612} are domain-specific. Dialog Acts (DA) \cite{stolcke-etal-2000-dialogue} are more formalized and generalized versions of intents. The international standard for DA annotations \cite{bunt-etal-2010-towards,bunt-etal-2012-iso,bunt-etal-2016-dialogbank} defined the concept of communicative functions in a dialog act. However, these functions are defined for a wide range of use cases. We note that a very restrictive and reworked subset of these suffices for our use cases. We believe the other features in the annotation scheme are novel or have an expanded range of possible values. 

The Global-Local model draws inspiration from the \citet{leeEndtoendNeuralCoreference2017} model for end-to-end neural coreference resolution. Like us, they have regions on interest embedded in a larger context. However, our models differ in several key ways: their span representation is a hand-crafted combination of token features while ours is a learned pooling of token representations. Also, their model is restricted to operating on contiguous spans (possibly due to unavailability of spans a priori, or that non-contiguous spans would lead to a combinatorial explosion), while our model has no such restriction. 

\section{Conclusion}

Improvements in the complexity of conversations that a dialogue system can handle have put tremendous pressure on NLU systems to capture fine-grained and domain-specific information. Difficulty in the data generation process means the ability to share data across clients is critical. We define intent features, a core set of general annotations, on intents that provide context and clarity on the exact nature of the user requests, and allow for a more natural and intelligent response from the dialogue manager. A NLU system that produces these intent features has been deployed in a production system with a dialogue manager that makes use of them.

To extract these intent features from an utterance, we propose a new neural network architecture, the Global-Local model, that fuses the representation of the content of a span of text and its global context through learned pooling functions. This model shows large improvements over several strong baselines. 

\section{Ethical Considerations}

The largest ethical concern about our work stems from our goal to share these intent features, and the models that identify them, across clients. It is critical to ensure that models trained for one client do not leak private user information to other clients. Given that our model is a simple classifier, opposed to a generative model, we do not believe information is leaking, but we are working on verifying this fact.

In addition to user privacy concerns, it is also important that our models do not underperform on a specific population of people. An internal tech report has investigated differences in performance based on user gender and has found none. This method will be applied to future models, as well as our currently deployed feature intent models, to make sure our models remain un-biased.

% Entries for the entire Anthology, followed by custom entries
\bibliography{anthology,custom}
\bibliographystyle{acl_natbib}

\end{document}